\documentclass[final]{clv2025}

\jname{preprint}
\jinfo{}
\renewcommand{\footmark}{}

\dochead{Short Paper}

\usepackage{amsmath}
\usepackage{booktabs}
\usepackage{graphicx}
\usepackage{tabularx}
\usepackage{array}
\usepackage{url}
\AtBeginDocument{\hypersetup{hidelinks}}
\usepackage{xcolor}

\runningtitle{Disagreement Structure}
\runningauthor{\"{O}hman, Koufakou \& Diegoli}

\begin{document}

\title{The Structure of Disagreement in Emotion Annotation}
\pagestyle{plain}
\author{Emily \"{O}hman\thanks{Corresponding author}$^{1}$, Anna Koufakou$^{2}$, Eugenia Diegoli$^{3}$}

\affilblock{
    \affil{Waseda University, Japan\\\quad \email{ohman@waseda.jp}}
    \affil{Florida Gulf Coast University, USA}
    \affil{Ca' Foscari University of Venice, Italy}
}
\maketitle

\begin{abstract}

Calculating inter-annotator agreement is an established practice in NLP, and disagreement-aware work increasingly retains the distribution of human judgments rather than reducing annotations to a single gold label. Keeping the distribution, however, does not explain why annotators disagree. This paper examines the types of disagreement that occur in emotion annotation, starting from multilabel annotations of 1,677 sentence-level units from Ernest Hemingway's \textit{The Old Man and the Sea} and testing the patterns they reveal against individual-level annotations from BRIGHTER and GoEmotions. Across the three datasets, we find three main recurring disagreement structures. Emotion categories overlap systematically; annotators differ both in how broadly they apply labels and in how they treat particular category boundaries, while sparse participation and explicit abstention can look like disagreement. Where intensity is rated, disagreement over whether an emotion is present can be separated from disagreement over its intensity once recognized. Similar aggregate disagreement can therefore arise from qualitatively different annotation processes. Hard aggregation removes most of this structure, while annotator vote-share representations keep marginal label frequencies but not the co-occurrence, annotator-specific, or procedural patterns that produced them. We argue that disagreement should be understood in terms of its underlying patterns before it is summarized, aggregated, or used as a modeling target, and propose a vocabulary for disagreement structure.

\end{abstract}


\section{Introduction}

Emotion annotation, in which annotators are asked to identify the emotion(s) expressed in a text, has received considerable attention in NLP and neighboring disciplines alike \cite{mohammad2022ethics, muhammad2025brighter, ohman2020emotion, bostanGoodNewsEveryoneCorpusNews2020}. Emotion annotation, however, is also a good place to observe a general problem in NLP: turning interpretation into data means deciding what matters and how to deal with competing interpretations. NLP pipelines have long treated inter-annotator agreement as a proxy for annotation quality and disagreement as something to be resolved by adjudication, majority voting, or filtering. That is convenient when the goal is a single target for training and evaluating a classifier. It is harder to defend when the object being annotated is highly context dependent and inherently open to subjective interpretation.

Perceived emotion is precisely such a case. Emotion in text cannot always be inferred from lexical content alone. This is especially visible in literary writing, where affect is communicated through narration, characterization, expectation, irony, and accumulated context, so the same passage often supports several plausible readings. The problem is not restricted to literary text, however. Even when the lexical content seems to point towards one single interpretation, for example because there are explicit emotion labels that guide the annotation, contextual information will drive the interpretation of specific elements, such as intensity or valence. It is possible to imagine situations in which supposedly negative emotion will be positive (e.g. anger in contexts of abuse), and supposedly positive emotion will be negative (e.g. unrequited love) \citep[][p. 46]{galasinski2004men}, with consequences in terms of inferred meanings. \citet{plank2022problem} makes the general argument that human label variation should not be assumed to be error, and perspectivist work in NLP has pushed for keeping multiple judgments rather than assuming a single ground truth \citep{aroyo2015truth,basile2021consider,cabitza2023perspectivist,leonardelli2023semeval,leonardelli2025lewadi,frenda2025perspectivist}. Emotion annotation fits this argument well, because categorical emotion schemes discretize something that is overlapping, graded, and interpretation-dependent \citep{cowen2021semantic,koufakou2025review}.


In what follows, we use literary annotation as the motivating case for retaining disagreement, because its interpretive character is particularly explicit. We then ask whether related disagreement structures recur in larger emotion resources collected under different annotation designs. Ernest Hemingway's \textit{The Old Man and the Sea} (TOMATS) therefore provides the initial case from which the distinctions are developed, while BRIGHTER \citep{muhammad2025brighter} and GoEmotions \cite{demszky2020goemotions} allow us to test and refine them across independently collected annotation data.

We build on prior work, which compared hard targets with vote-share and intensity-weighted soft targets and examined downstream behavior \citep{ohman2026quality}. The findings indicated that hard labels can outperform soft targets on thresholded F1, while the predicted probabilities of soft targets are better aligned with the spread of human judgments. What counts as a better model depends on what is being optimized: evaluation metrics or generalizability. Here we ask the question that comes before that one: what kind of disagreement are we looking at?

The same amount of measured disagreement can come from very different processes. In emotion annotation tasks where annotators are given a predefined list of emotion labels to choose from, two annotators may pick related emotions because the passage genuinely carries both. Or, they may interpret the same label differently and draw the boundary between distinct emotion categories in a different place. Or one of them may simply have left the item blank. Annotators may also disagree over whether an emotion is present while agreeing relatively closely about its intensity once recognized, or agree on its presence while differing more substantially in intensity. A single agreement score cannot distinguish these cases, and neither can an aggregated probability vector, as we show below.

Using the TOMATS annotations first analyzed in \citet{ohman2026quality} (see Section~\ref{ssec:tomats}), we identify three observable disagreement signatures:

\begin{itemize}
\item semantic overlap, where annotators select different labels for the same passage, but certain pairs of labels recur as alternative readings across annotators;

\item boundary structure, where annotators differ systematically in which emotions they are willing to assign together; and

\item procedural sparsity, where missing or unusually thin annotation produces apparent disagreement that is not a competing reading at all.
\end{itemize}

We then use BRIGHTER and GoEmotions to ask which of these signatures recur under different annotation designs and where the larger datasets require the picture to be refined. Despite substantial differences in annotation design, label inventories, and annotator populations, related disagreement structures emerge across the three resources. GoEmotions separates general annotation breadth from pair-specific boundary behavior, while BRIGHTER allows disagreement over emotion presence to be distinguished from disagreement over intensity once an emotion is recognized. 

We say \textit{signature} rather than \textit{cause} on purpose. All analyses are observational. Cross-dataset recurrence makes the patterns less likely to be idiosyncratic to TOMATS, but it does not establish why any individual disagreement occurred. These are patterns we can see in the annotation record that aggregation would erase.

\section{Disagreement, Interpretation, and Aggregation}

\subsection{From agreement to perspectivism}

Inter-annotator agreement has traditionally been one of the main diagnostics of annotation reliability. A recent large-scale analysis of NLP dataset construction found that agreement is widely reported but often insufficiently interpreted, while majority voting remains the dominant adjudication strategy \citep{klie2024analyzing}. Cohen's kappa and Krippendorff's alpha correct observed agreement for chance, but interpreting the resulting number depends heavily on the structure of the task and the label distribution \citep{artstein2008survey}. Multilabel annotation complicates this further, because several categories can be valid at once and apparent non-agreement is often partial overlap rather than a contradiction. In TOMATS, the per-emotion Krippendorff's alpha values fall between 0.09 (trust) and 0.28 (sadness), which by conventional guidelines would be dismissed as unusable \citep{ohman2026quality}. We report them here for the same reason they were reported in that paper: they are a familiar summary of annotation reliability, yet they are difficult to interpret meaningfully for a multilabel emotion task.

Agreement statistics do not measure a single underlying quantity. Different measures target different aspects of the annotation process and rely on different assumptions about the scale and structure of the judgments \citep{artstein2008survey,hallgren2012computing,gwet2014handbook,james2026counting}. 
Agreement is therefore better understood as a property of the interaction between the annotation target, response format, rater design, and summary statistic than as a single property of a dataset.

Recent perspectivist work has emphasized that disagreement may arise for different reasons and that aggregation can discard meaningful information \citep{fleisig2024perspectivist}. What remains less clear is how the structure of disagreement can be diagnosed from annotation records rather than treated collectively as ``human label variation.'' Underlying this measurement problem is a conceptual one. High agreement does not necessarily mean the annotators have captured the phenomenon more faithfully, and low agreement does not necessarily mean the annotation is bad. \citet{plank2022problem} calls this the problem of human label variation: our evaluation practices assume a correct label exists even when human responses vary systematically. Individual variability tends to be particularly high for subjective and private experiences like emotions, which can be difficult to express in propositional terms and are often fundamentally ambiguous or vague \citep{wharton2023pragmatics}. Perspectivist NLP treats that variation as information. \citet{basile2021consider} argue that disagreement can originate in the annotator, the data, or the context, and that evaluation should reflect this rather than remove it. The ``Learning with Disagreements'' shared tasks ask systems to model distributions of judgments or individual annotators instead of aggregated labels \citep{leonardelli2023semeval,leonardelli2025lewadi}, and soft-label and disaggregated-label approaches have been studied extensively in recent work \citep{uma2021learning,kurniawan-etal-2026-training}. A small but growing number of datasets now release annotator-level labels rather than only the aggregate \citep{muhammad2025brighter,demszky2020goemotions,barz2025understanding}. Two of these resources are particularly useful for the present analysis because they preserve different aspects of the annotation process: BRIGHTER provides individual ordinal ratings for each emotion dimension, while GoEmotions preserves persistent rater identifiers together with fine-grained multilabel choices. They therefore allow different parts of disagreement structure to be examined rather than only its aggregate magnitude.

\subsection{Emotion annotation as interpretive data}

Emotion makes these issues especially visible. Most NLP resources operationalize emotion through a fixed set of categories, often derived from psychological models such as \citet{plutchik1984emotions}. Categories make computation possible, but affective-science research likewise finds that emotional states often occupy graded and blended regions rather than falling into sharply separated classes \citep{cowen2021semantic}. This variability gets lost in majority voting.

The interpretive problem is particularly visible in literary annotation, in which the annotator is interpreting a represented situation rather than reporting a felt one, a task that demands greater mental effort and is inherently more subjective \citep{thomas2022crowd}. The annotation process presents specific challenges because emotion can be encoded in text more or less overtly and at different levels. A sentence may contain an explicit emotion word, imply a character's state, provoke an emotive response in the reader, or only acquire emotional weight from where it sits in the narrative. Annotation guidelines have to decide which of these counts, and the guidelines rarely say so explicitly. This makes TOMATS a useful motivating case, but the underlying measurement problem is broader: emotion annotations in non-literary datasets also depend on task instructions, category definitions, context, and judgments about what counts as evidence for an emotion.

Turning affect into data is therefore not a matter of recording something that was already there. The taxonomy, the amount of context, the instructions, the annotation interface, and the aggregation method all shape the measurement \citep{laaksonen2023datafication}. That does not make computational emotion analysis impossible, but it does mean that the annotation process needs to be documented rather than assumed away \citep{mohammad2022ethics}.

So we move from ``how much do the annotators agree?'' to ``what does their disagreement look like?''. Agreement coefficients and vote distributions are still useful summaries. They are not a substitute for looking at the annotation patterns they were computed from.

\section{Data and Analytic Strategy}

We use three emotion-annotation resources with complementary designs (see Table~\ref{tab:datasets}). TOMATS provides the motivating case and supports analyses of semantic overlap, annotator-specific co-occurrence patterns, and procedural sparsity. BRIGHTER provides complete ordinal ratings for multiple emotion dimensions and therefore allows disagreement over emotion presence to be separated from disagreement over intensity. GoEmotions preserves persistent rater identifiers and fine-grained multilabel judgments, allowing general annotator response style to be distinguished from pair-specific category-boundary behavior. We do not pool the three resources or compare their absolute disagreement levels. Instead, we use their different annotation designs to ask whether particular disagreement structures can be observed independently across datasets.

\begin{table*}[htbp]
\centering
\caption{Overview of the three emotion-annotation resources used in this study}
\label{tab:datasets}
\small
\begin{tabularx}{\textwidth}{
>{\raggedright\arraybackslash}p{0.11\linewidth}
>{\raggedright\arraybackslash}p{0.10\linewidth}
>{\raggedright\arraybackslash}p{0.09\linewidth}
>{\raggedright\arraybackslash}p{0.10\linewidth}
>{\raggedright\arraybackslash}p{0.20\linewidth}
>{\raggedright\arraybackslash}X}
\toprule
Dataset & Size & Emotions & Annotators / raters & Annotation format  & Role in this study \\
\midrule
TOMATS \citep{ohman2026quality}
& 1,677 text units
& 8 core emotions
& 3 annotators
& Up to three labels per item; contextual literary annotation; overall intensity field; explicit no-emotion option
& Motivating case for semantic overlap, boundary structure, and procedural sparsity \\

BRIGHTER \citep{muhammad2025brighter}
& 5,655 text units in English
& 6 dimensions
& 5--30 ratings per item
& Individual 0--3 ordinal rating on every emotion dimension
& Tests structured co-occurrence and separates presence from conditional-intensity disagreement \\

GoEmotions \citep{demszky2020goemotions}
& 211,225 individual annotation rows
& 27 emotions + neutral
& 82 raters
& Fine-grained multilabel annotation with persistent rater IDs
& Tests general response style, pair-specific boundaries, and procedural/abstention behavior \\
\bottomrule
\end{tabularx}
\end{table*}

\subsection{The TOMATS annotations}
\label{ssec:tomats}

The data are 1,677 roughly sentence-level units from \textit{The Old Man and the Sea}, annotated in English by three research assistants using an expanded version of Plutchik's eight-category scheme: anger, anticipation, disgust, fear, joy, sadness, surprise, and trust.

The annotators read the whole text first and then annotated the units in order, in context. They could assign up to three emotions per unit. The annotation protocol also included an overall 0--10 intensity judgment for the passage, together with an explicit no emotion option and the possibility of adding a free-text emotion or note where the eight categories did not fit. Because the TOMATS intensity value is not cleanly tied to each individual emotion label, it is not used for the presence-versus-intensity analysis in this article. The free-text labels are also outside the present analysis, although as we note in Section~\ref{sec:limitations} they may themselves reflect an additional form of disagreement.

The three label fields were called primary, secondary, and tertiary. These are just the first, second, and third labels entered for an item. They do not mean that the first emotion is psychologically primary or more important than the others.

Coverage differed substantially between annotators (Table~\ref{tab:1}). A1 supplied a primary emotion for 1,593 items, plus 1,258 secondary and 443 tertiary labels. A3 supplied primary labels for 1,369 items, plus 1,200 secondary and 566 tertiary. A2 supplied a primary emotion for 184 items and added only 13 secondary and 3 tertiary labels. Blanks have to be kept separate from the explicit no emotion option: an empty field is not a judgment that there is no emotion present.

\begin{table}[t]
\centering
\caption{Annotation coverage by annotator. Reproduced from \citep{ohman2026quality}}
\label{tab:1}
\begin{tabular}{lrrr}
\toprule
Annotator & Items with primary emotion & Secondary labels & Tertiary labels \\
\midrule
A1 & 1,593 & 1,258 & 443 \\
A2 & 184 & 13 & 3 \\
A3 & 1,369 & 1,200 & 566 \\
\bottomrule
\end{tabular}
\end{table}

The TOMATS annotations were introduced in \citet{ohman2026quality}, which focused primarily on alternative aggregation strategies and downstream classification. The present analysis returns to the annotator-level records with a different aim: to formalize distinct disagreement signatures, quantify them more precisely, and test whether related structures recur in independently collected emotion-annotation datasets.

\subsection{Cross-dataset resources}

Table~\ref{tab:datasets} summarizes the three resources. We do not pool them or compare their absolute disagreement levels; each is used for the analyses its design supports.

The English individual-annotation release of BRIGHTER contains 5,655 items (referred to in the original paper as ``text snippets''), with between 5 and 30 ratings per item across six emotion dimensions, named in the raw pre-release category set as anger, fear/anxiety, happiness, sadness, social warmth, and surprise \citep{muhammad2025brighter}. Each emotion is rated on a 0--3 ordinal scale. This design allows us to distinguish disagreement about whether an emotion is present at all from disagreement about its intensity among annotators who judge it to be present.

The raw GoEmotions annotations contain 211,225 individual annotation rows from 82 raters and preserve rater identifiers across items \citep{demszky2020goemotions}. Annotators could select multiple fine-grained emotion labels, and the released data also contain an explicit \texttt{example\_very\_unclear} response. Persistent rater identifiers allow us to compare an annotator's behavior with that of other annotators who judged the same items 
and to distinguish general differences in annotation breadth from pair-specific differences in how neighboring emotion categories are used.

\subsection{TOMATS disagreement diagnostics}

We use three diagnostics to characterize the disagreement visible in TOMATS.

The first is cross-annotator emotion association: when A1 selects a given emotion, how often A3 selects each emotion on the same item, relative to A3's overall rate for that emotion (lift). Because both can supply up to three labels, one item contributes several emotion pairs. The matrix shows which pairs of choices recur, not which reading is correct.

The second is within-annotator co-occurrence: which categories a single annotator assigns together to the same item. We report it as lift relative to the annotator's own marginal label frequencies, restricted to items on which the annotator supplied at least one core emotion.

The third is the full versus filtered pairwise agreement comparison. Agreement is computed independently for each of the eight emotions and summarized as macro- and micro-F1. The full condition uses the whole corpus. The filtered condition keeps only the 184 items where A2 supplied at least one core emotion label. The gap between the two shows how much of the apparent disagreement is associated with differences in annotation coverage.

Table~\ref{tab:2} summarizes the three TOMATS-derived disagreement signatures that motivate the cross-dataset analyses.

\begin{table*}[t]
\centering
\caption{Three disagreement signatures identified in the TOMATS annotations}
\label{tab:2}
\small
\renewcommand{\arraystretch}{1.4}
\begin{tabularx}{\textwidth}{>{\raggedright\arraybackslash}p{0.13\linewidth}>{\raggedright\arraybackslash}X>{\raggedright\arraybackslash}X>{\raggedright\arraybackslash}X}
\toprule
Signature & Observable pattern & Plausible interpretation & What aggregation obscures \\
\midrule
Semantic overlap & Recurrent cross-annotator emotion pairings & Multiple or blended readings; overlap in the taxonomy & Which alternatives regularly compete or co-occur \\
Boundary structure & Different within-annotator co-occurrence patterns & Different implicit category boundaries or labeling strategies & Which annotators treat which categories as compatible \\
Procedural sparsity & High rates of blank or thin annotation & Nonresponse, limited protocol engagement, or a restrictive labeling style & Whether an absence represents disagreement or lack of engagement \\
\bottomrule
\end{tabularx}
\end{table*}

\subsection{Cross-dataset disagreement diagnostics}

The BRIGHTER analyses address two questions. First, we examine whether emotion categories show structured co-occurrence rather than independent presence. An emotion is treated as present when its rating is greater than zero. For each pair of emotions, we calculate co-occurrence lift as the observed rate of joint presence within an individual judgment divided by the rate expected from the two marginal frequencies. Values above one indicate that two emotions co-occur more often than expected under independence.

Second, we distinguish disagreement over emotion presence from disagreement over conditional intensity. Presence disagreement is measured as binary entropy over ratings of zero versus greater than zero. Conditional-intensity disagreement is calculated only among annotators who assign a positive rating and is measured as the mean pairwise absolute distance between their intensity ratings, normalized to the positive portion of the scale. To avoid unstable estimates, this analysis is restricted to item--emotion combinations with at least three positive ratings.

The GoEmotions analyses address annotator-specific response style and category-boundary behavior. For each rater, we first estimate item-adjusted label cardinality. On each item, the number of labels selected by the focal rater is compared with the mean number selected by the other clear raters who annotated the same item (that is, raters who did not mark the item as \texttt{example\_very\_unclear}), and these differences are then averaged across items. Negative values therefore indicate a relatively restrictive annotator and positive values a relatively expansive annotator. Analyses of overall response style are restricted to raters with at least 100 matched-item comparisons.

We then examine six neighboring emotion pairs: anger--annoyance, fear--nervousness, joy--excitement, sadness--grief, gratitude--admiration, and surprise--realization. For each pair, we estimate whether a rater tends to select both categories together more or less often than co-raters judging the same items. Because a generally expansive annotator will mechanically be more likely to select both labels, pair-specific merging tendency is residualized against that rater's overall item-adjusted label cardinality. The resulting residual therefore captures pair-specific boundary behavior beyond general annotation breadth.

Finally, we examine the use of GoEmotions' explicit \texttt{example\_very\_unclear} response as a procedural comparison, computed for each rater as the proportion of their annotation rows marked very unclear. This is not equivalent to the sparse behavior observed in TOMATS, but it provides an independently recorded case in which an absence of emotion labels does not straightforwardly mean that the annotator judged every emotion to be absent.

\section{Results}

\subsection{Semantic overlap is structured}
\label{ssec:overlap}

Annotators A1 and A3 give the most informative comparison, since both annotate most of the corpus and utilize the secondary and tertiary labels. They choose the same primary emotion for 42\% of items, and share at least one emotion for 62\% \citep{ohman2026quality}. That is, a large share of what looks like primary-label disagreement is actually partial multilabel agreement. 

\begin{figure}[t]
\centering
\includegraphics[width=0.8\linewidth]{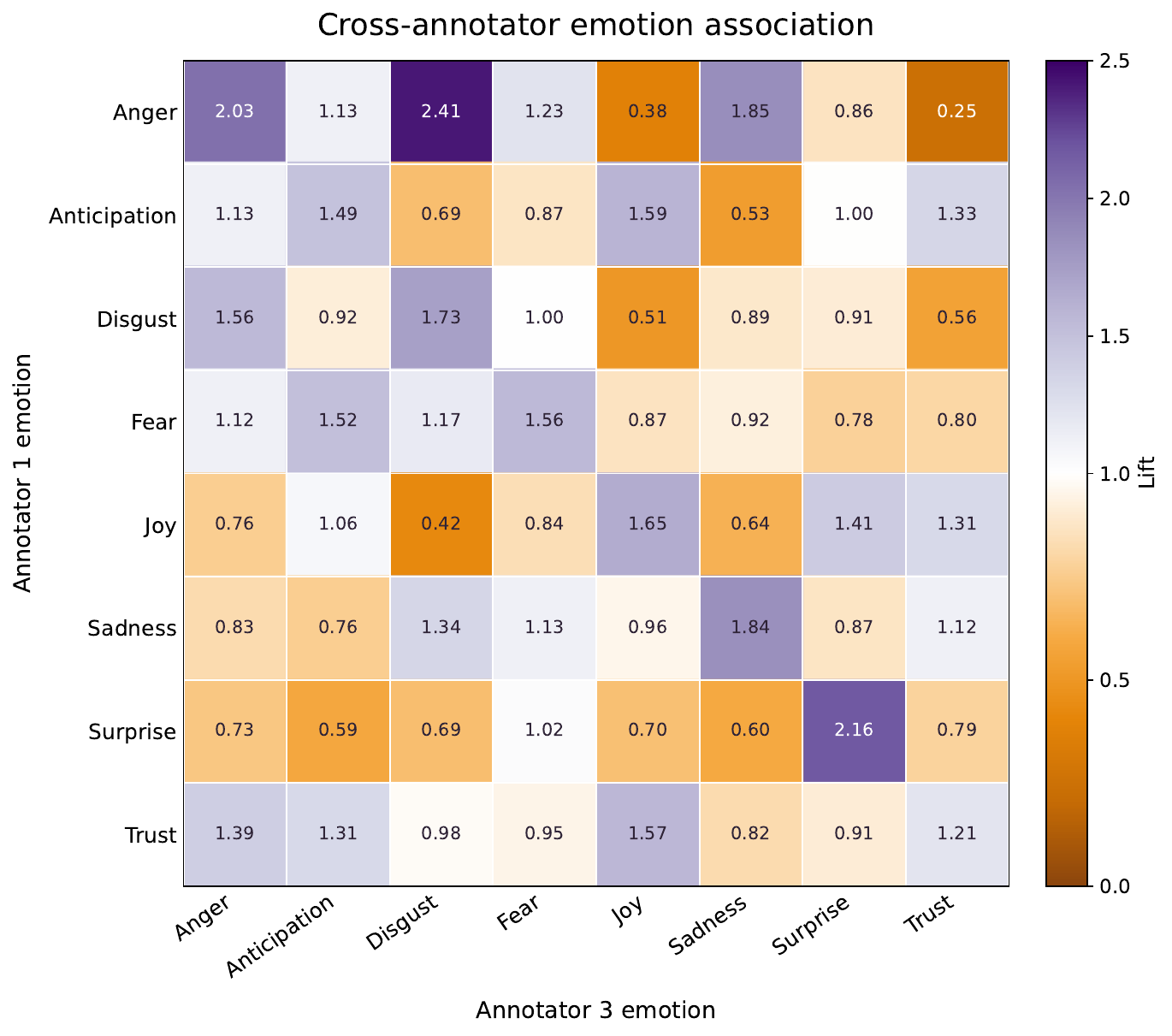}
\caption{Cross-annotator emotion association between A1 and A3, expressed as lift. Each cell shows
$P(\mathrm{A3\ includes\ column} \mid \mathrm{A1\ includes\ row}) / P(\mathrm{A3\ includes\ column})$.
Values above 1 indicate that A3 selected the column emotion more often than expected when A1 selected the row emotion; values below 1 indicate less frequent co-selection than expected. Cell labels report raw lift values, while color intensity is based on log$_2$(lift) and centered at 1.}
\label{fig:cross-overlap}
\end{figure}

To examine the structure of the remaining divergence, we calculate cross-annotator lift (Figure~\ref{fig:cross-overlap}). For each emotion selected by A1, we measure how often A3 also selects each emotion, relative to A3's overall base rate for that category. A lift value of 1 therefore indicates no association beyond chance expectation given A3's marginal label frequency, values above 1 indicate enrichment, and values below 1 indicate depletion. The heatmap is colored on a log$_2$-transformed scale centered at lift = 1, while the cell labels report the original lift values.

Anger--disgust stands out most clearly. When A1 selects anger, A3 selects disgust far more often than A3's overall use of disgust would predict (lift = 2.41), and the association also holds in the other direction (1.56). Fear--anticipation is weaker and runs one way: when A1 selects fear, A3 selects anticipation more often than expected (1.52), but when A1 selects anticipation, A3 is slightly less likely than usual to choose fear (0.87). Several other cells reach similar values (anger--sadness 1.85, anticipation--sadness 1.59, trust--joy 1.57), so fear--anticipation is not exceptional on this measure. We single it out because it sits awkwardly in Plutchik's geometry, as discussed below. Same-emotion associations are also strong for several categories, including surprise, anger, sadness, and disgust. Disagreement is structured around particular category relations rather than spread evenly across the taxonomy.

The anger--disgust and fear--anticipation relationships are consistent with patterns observed in our earlier emotion-arc work on a much larger corpus \citep{ohman2024emotionarcs}, suggesting that these associations are not simply an artifact of this text or these three annotators. More broadly, recent work shows that emotion labels should not be treated as fixed, mutually exclusive units: multilabel judgments frequently combine emotions associated with different coarse polarities \citep{ohman2026mixed,larsen2014case}, and even same-named categories such as \emph{fear} can behave differently across datasets and annotation schemes \citep{ohman2026fear}. The recurrent associations observed here are therefore better interpreted as evidence of structured category relations and annotator-specific boundary placement than as arbitrary label confusion.

Anger and disgust are the easy case. They sit next to each other on Plutchik's wheel; they combine into the primary dyad contempt \citep{plutchik1984emotions}, and both plausibly describe a hostile or aversive passage. Their recurrent overlap suggests that forcing a choice between them draws the distinction more sharply than annotators themselves do when inferring emotion from text.

Fear and anticipation are the more interesting pair, because they are not neighbors. They are three petals apart on the wheel and form the tertiary dyad anxiety \citep{plutchik1984emotions}. In narrative prose, anticipated danger can plausibly be read as either expectation or fear, and the within-annotator results below suggest this is how the annotators treated it: as alternative readings of the same passage rather than as two emotions present at once. This points to a gap between the taxonomy's conceptual geometry and the affective relationships instantiated in the text. An overlap matrix can reveal that gap; an agreement coefficient cannot.

The cross-annotator and within-annotator analyses capture different kinds of relationship between categories. In the within-annotator lift analysis, restricted to items on which the annotator supplied at least one core emotion label, anger--disgust is strongly associated for A1 (lift = 2.31) and also occurs above independence for A3 (lift = 1.24). Other positively associated pairs differ between the two annotators: anticipation--joy and joy--surprise both reach a lift of 1.42 for A1, whereas A3's strongest associations are disgust--sadness (1.49), anger--anticipation (1.40), and anger--joy (1.31). The within-annotator structure is therefore systematic but annotator-specific.

The contrast with fear--anticipation is particularly informative. Although A1's fear is associated with A3's anticipation, the two emotions occur less often than expected under independence within both A1 (lift = 0.88) and A3 (lift = 0.80). In other words, the pair appears more often as an alternative categorization across annotators than as a combination assigned within a single judgment. Cross-annotator overlap and within-annotator co-occurrence therefore capture different aspects of disagreement structure.

\begin{figure}[t]
\centering
\includegraphics[width=0.6\linewidth]{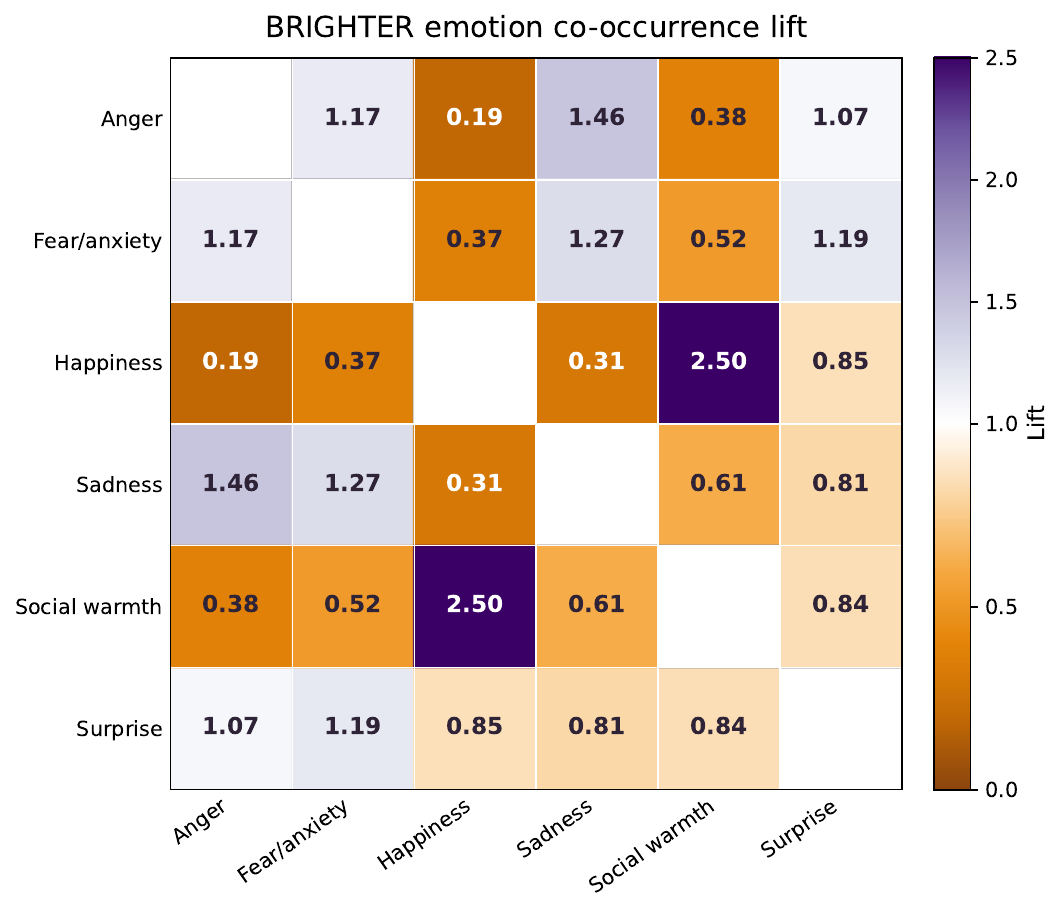}
\caption{Structured co-occurrence among BRIGHTER English emotion categories. Each cell reports co-occurrence lift within individual judgments. Values above 1 indicate more joint presence than expected from the marginal frequencies of the two emotions, values below 1 indicate less joint presence than expected, and diagonal cells are omitted.}
\label{fig:brighter-lift}
\end{figure}

BRIGHTER provides a directly comparable within-judgment measure. Its label inventory differs from TOMATS, so we do not expect the same category pairs, but its categories likewise show substantial departures from independence (Figure~\ref{fig:brighter-lift}). Happiness--social warmth has the strongest positive lift (2.50), followed by anger--sadness (1.46), fear/anxiety--sadness (1.27), fear/anxiety--surprise (1.19), and anger--fear/anxiety (1.17). By contrast, anger--happiness (0.19) and happiness--sadness (0.31) occur much less often than expected from their marginal frequencies. Across the two resources, emotion labels therefore participate in structured compatibility relations, although the particular relations depend on the inventory, task, and data.

\subsection{Annotators differ in response style and category boundaries}
\label{ssec:boundaries}

Cross-annotator matrices show which labels differ between annotators. Within-annotator matrices show something else: which labels each annotator tends to combine within the same judgment. 

\begin{figure*}[t]
\centering

\begin{minipage}[t]{0.42\textwidth}
    \centering
    \includegraphics[width=\linewidth]{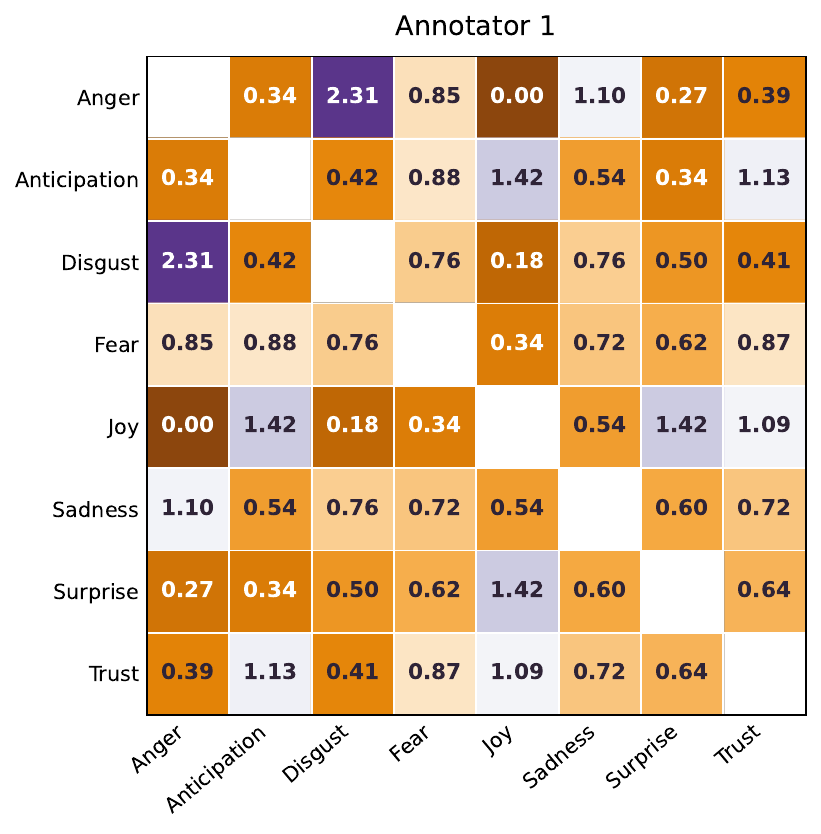}
    \textbf{(a)} Annotator 1
\end{minipage}
\hfill
\begin{minipage}[t]{0.54\textwidth}
    \centering
    \includegraphics[width=\linewidth]{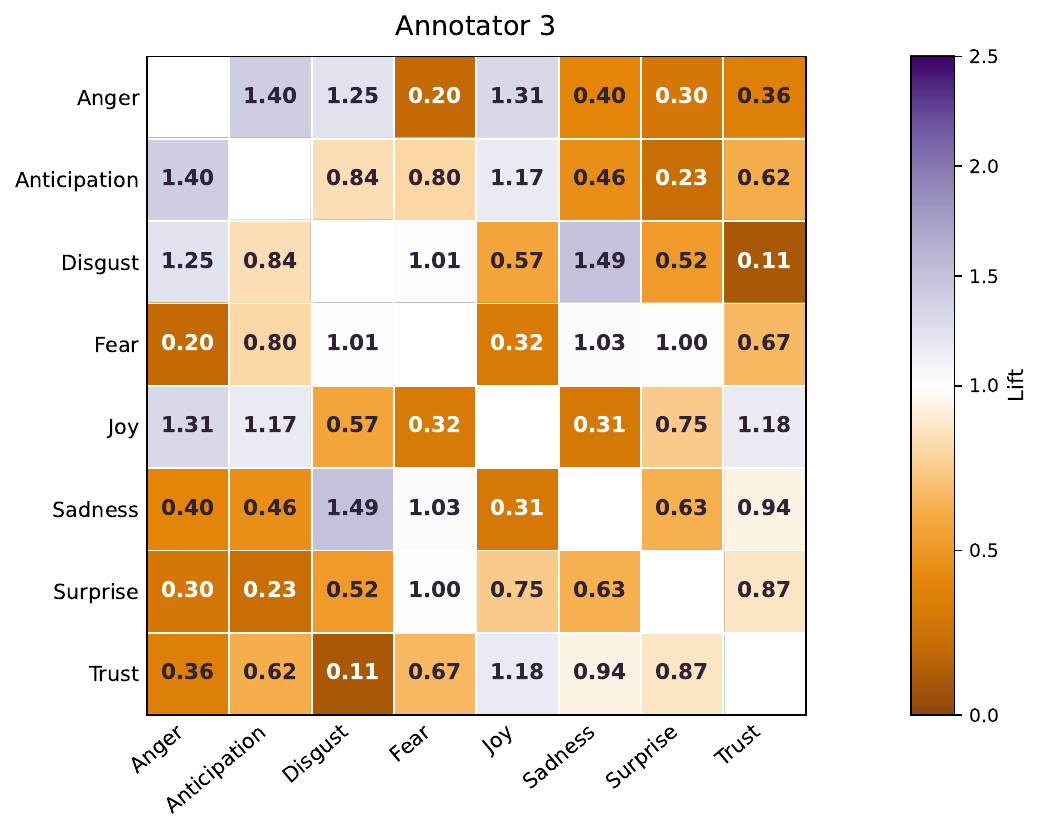}
    \textbf{(b)} Annotator 3
\end{minipage}

\caption{Within-annotator emotion co-occurrence lift for Annotators 1 and 3. Values above 1 indicate that a pair of emotions was selected together more often than expected from the annotator's marginal use of the two categories; values below 1 indicate less co-occurrence than expected. Diagonal cells are omitted. Cell labels report raw lift values, and the color scale is centered at lift = 1.}
\label{fig:tomats-within-lift}
\end{figure*}

Both matrices (Figure~\ref{fig:tomats-within-lift}) show clear internal structure. Anticipation, joy, and trust cluster in both, and anger and disgust co-occur in both. Fear and sadness connect to several other regions of the label space. Neither annotator is combining categories at random.

The two structures are not the same, though. A1 keeps sharper boundaries around some mixed-valence combinations, while A3 allows broader overlap. Anger--joy is the clearest example: it does not occur in A1's matrix at all, but is positively associated for A3 (lift = 1.31). More generally, A3 spreads co-occurrence across a wider range of pairs. In TOMATS alone, however, this broader pattern cannot tell us whether A3 draws particular semantic boundaries differently or simply applies more labels overall. A3 does use somewhat broader multilabel annotations: among items containing at least one core emotion label, A3 assigns an average of 2.21 core labels per item compared with 2.06 for A1. The difference in their co-occurrence structures may therefore combine general annotation breadth with pair-specific category use.

This suggests boundary structure rather than disagreement in the usual sense, because it is not necessarily a case of A1 seeing emotion A where A3 sees emotion B. The two may differ in their conceptualizations of emotions A and B in the first place, and hence in whether A and B can describe the same passage and whether they can co-occur with other emotions. An annotator who treats the categories as fairly discrete will produce fewer mixed combinations than one who treats emotional states as layered, without either necessarily being wrong about the passage.

This matters for how multilabel annotation is read. If one annotator selects anger and disgust together and another selects anger alone, the second has not necessarily rejected disgust. They may have slightly different understandings of what anger and disgust entail, they may be applying a higher threshold for adding a label, or a narrower idea of what counts as simultaneous emotional content. The resulting label set encodes an interpretation of the passage and an interpretation of the task, and we cannot separate the two after the fact.

One alternative explanation cannot be ruled out with this design: priming or order effects. Annotation judgments are affected by preceding judgments and by item order \citep{mathur2017sequence,beck2024order}, and affectively related concepts facilitate each other in processing \citep{rossell2004semantic}, so an annotator who has already selected one emotion may find a related category more available for the second or third slot. Label order was not manipulated, so priming is a plausible mechanism here, not a demonstrated one. An aggregated target would not let us investigate it either: the annotator-specific co-occurrence structure needed to do so is exactly what disappears when only label frequencies are kept.

\begin{figure}[t]
\centering
\includegraphics[width=0.8\linewidth]{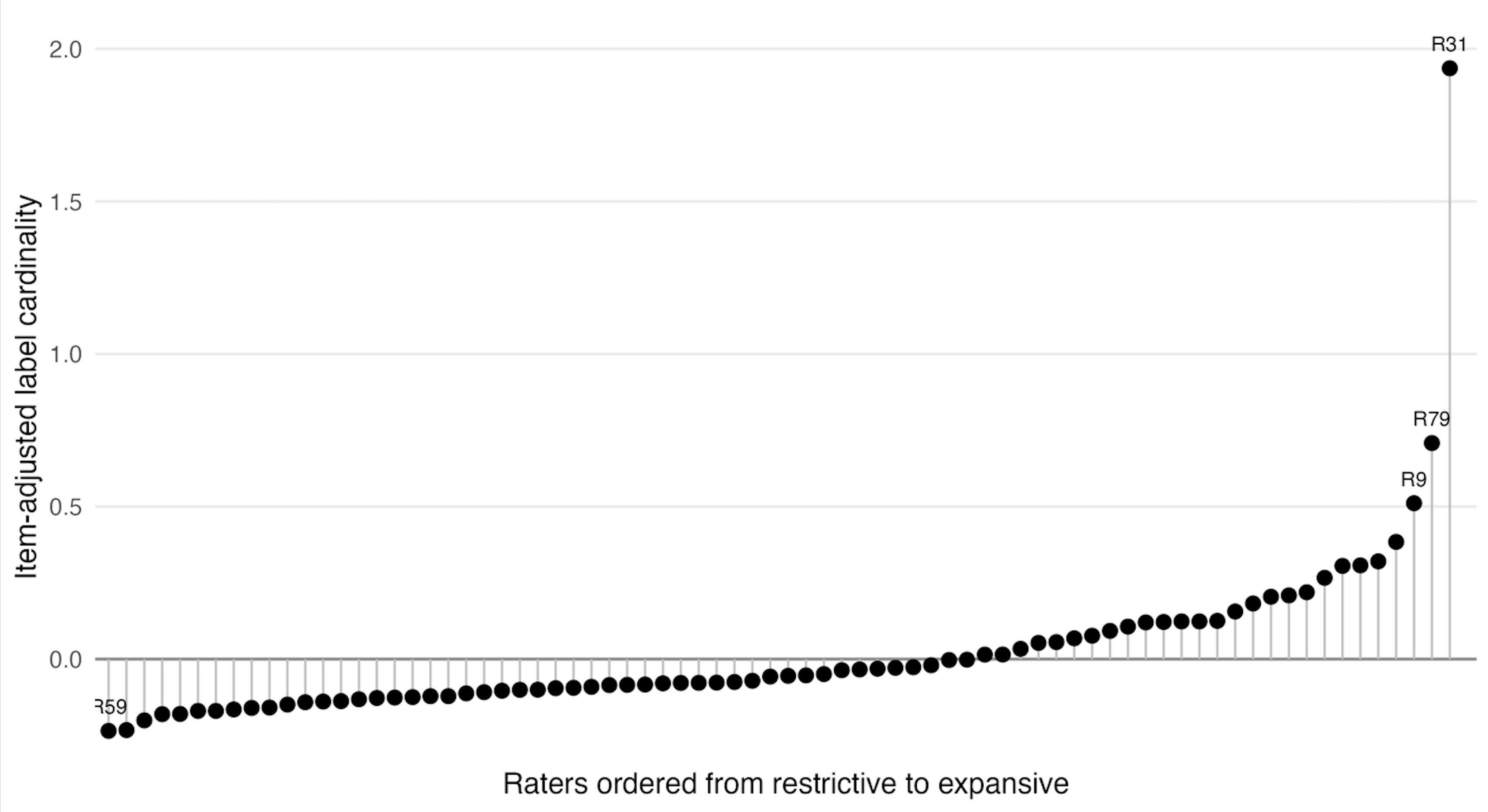}
\caption{GoEmotions: item-adjusted label cardinality by rater, ordered from most restrictive to most expansive.}
\label{fig:goemotions-breadth}
\end{figure}

GoEmotions allows general response style to be examined at a larger scale. Among the 76 raters with at least 100 matched-item comparisons, item-adjusted label cardinality ranges from approximately $-0.235$ to $+1.936$, with a median of $-0.056$ and an interquartile range of $-0.121$ to $+0.096$ (Figure~\ref{fig:goemotions-breadth}). The extreme upper value is contributed by a single rater (R31); excluding R31 leaves the center of the distribution essentially unchanged (median $=-0.057$, IQR $=-0.121$ to $+0.085$), although the observed maximum falls to $+0.708$. Negative values indicate that a rater tends to select fewer labels than co-raters judging the same items, while positive values indicate broader multilabel use. Most raters lie close to or slightly below the co-rater mean, while a long positive tail contains raters who select substantially more categories. Because the comparison is made within item, the differences cannot be attributed simply to some raters receiving more emotionally complex texts. GoEmotions therefore provides large-scale evidence that annotators differ in general annotation breadth.

\begin{figure}[t]
\centering
\includegraphics[width=0.95\linewidth]{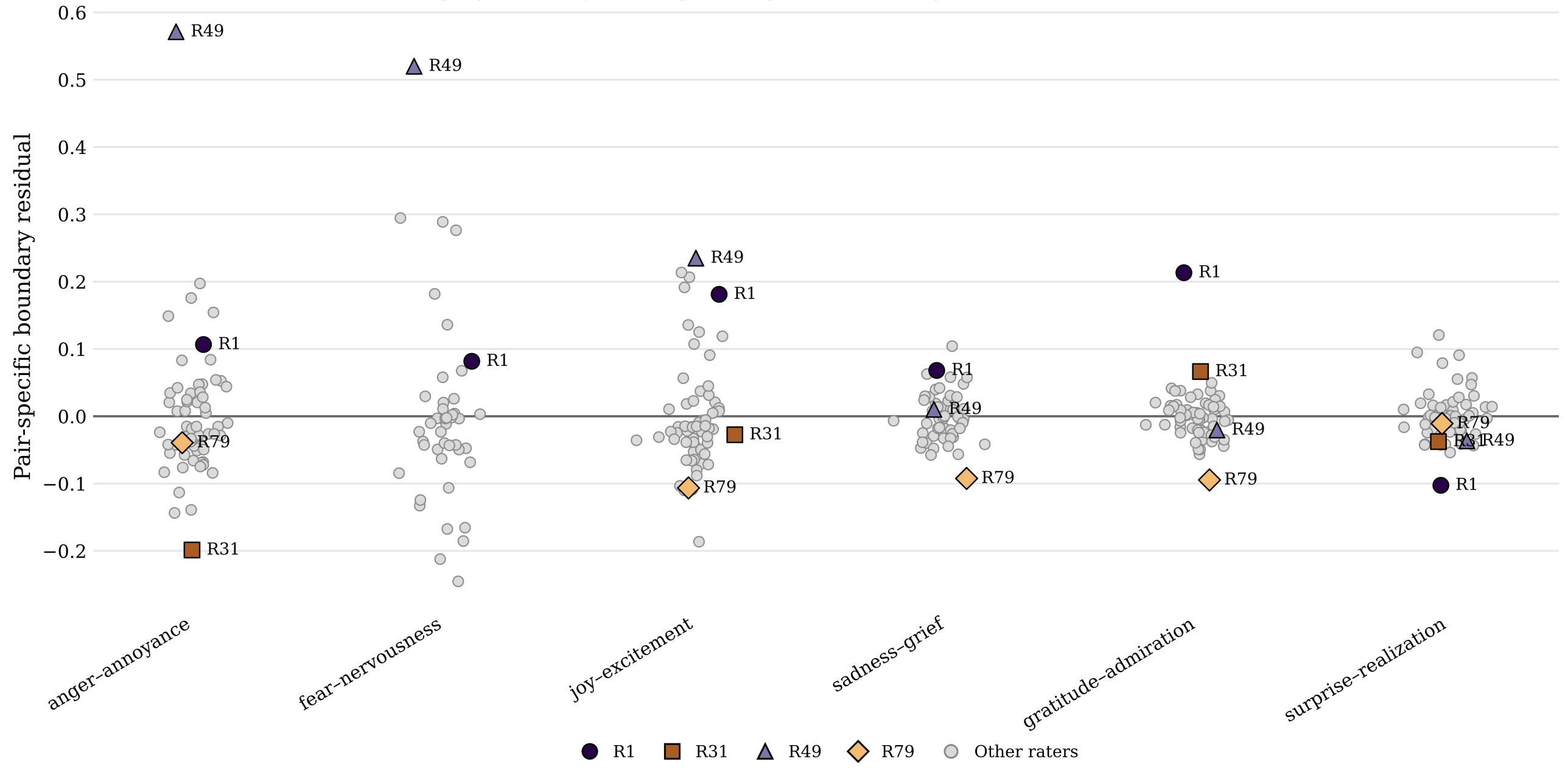}
\caption{GoEmotions: pair-specific semantic-boundary residuals for six neighboring emotion pairs, after controlling for each rater's overall annotation breadth.}
\label{fig:goemotions-boundaries}
\end{figure}

General breadth does not, however, fully explain how annotators treat particular neighboring categories. For six semantically related pairs, a residual above zero means that an annotator merges that pair more often than would be expected from their overall tendency to multilabel; a negative residual means that they keep the pair more distinct than expected. The analysis includes 64 raters for anger--annoyance, 46 for fear--nervousness, 61 for joy--excitement, 57 for sadness--grief, 67 for gratitude--admiration, and 62 for surprise--realization. Individual raters can be relatively merging for one pair while separating another (Figure~\ref{fig:goemotions-boundaries}). Semantic-boundary behavior is therefore not reducible to a single expansive--restrictive response style.

Annotator variation therefore has at least two observable levels: a general response style, which governs how readily someone multilabels, and local semantic-boundary behavior, which governs which neighboring categories they merge or separate beyond what their overall breadth alone would predict. The A1/A3 contrast in TOMATS plausibly mixes both, and we therefore read it as suggestive evidence of boundary structure rather than as a clean case of it. Neither level should be read as a stable personality trait. What the analysis shows is that these differences recur across items within the annotation record, which is exactly the kind of structure that disappears once annotator identifiers are dropped.

\subsection{Sparsity can look like disagreement}
\label{ssec:sparsity}

A2 shows the third signature. This annotator supplied a core emotion for 184 of 1,677 items and very rarely added a second or third label. The difference from A1 and A3 is large enough that simple pairwise comparisons mostly measure coverage.
Table~\ref{tab:3} reproduces the agreement values from \citet{ohman2026quality}.

\begin{table}[t]
\centering
\small
\caption{Pairwise multilabel agreement on the full TOMATS corpus and the A2-filtered subset}
\label{tab:3}
\begin{tabular}{lrrr}
\toprule
Pair & Condition & Macro-F1 & Micro-F1 \\
\midrule
A1--A2 & Full & 0.0788 & 0.0882 \\
A1--A2 & Filtered & 0.3949 & 0.4857 \\
A1--A3 & Full & 0.3794 & 0.3952 \\
A1--A3 & Filtered & 0.4899 & 0.5187 \\
A2--A3 & Full & 0.0829 & 0.0899 \\
A2--A3 & Filtered & 0.3797 & 0.4664 \\
\bottomrule
\end{tabular}
\end{table}

Restricting the comparison to the items A2 actually labeled raises A1--A2 and A2--A3 macro-F1 roughly fivefold (Table~\ref{tab:3}). A jump of that size cannot be read straightforwardly as evidence that A2 has a radically different interpretation of the text. A substantial part of the apparent disagreement is associated with the fact that A2 did not contribute labels to most items.

Filtering also raises A1--A3 macro-F1, from 0.3794 to 0.4899. So the filtered subset is not simply the full dataset minus A2's missing responses; it is a particular selection of passages. Whatever led A2 to annotate those 184 items also led them to select passages on which A1 and A3 agreed more strongly. Two things are therefore entangled: annotator sparsity and item selection. We cannot separate them with these data, and it is worth saying so rather than reporting the filtered numbers as corrected agreement values.

This sensitivity to filtering is not unique to the present data. Agreement estimates in emotion resources can change substantially when analyses are restricted to subsets selected on the basis of annotation behavior. For example, AffectNet reported substantially higher valence concordance after restricting analysis to images for which annotators had already agreed on the categorical emotion label \citep{mollahosseini2019affectnet}. Such estimates describe the retained subset rather than the original annotation problem, and should therefore not be interpreted as corrected versions of the unfiltered agreement value.

This makes the apparently simple quality-control decision---keep A2 or drop them---harder than it looks. In the earlier modeling experiments, excluding A2 lowered macro-F1 slightly under every training objective, and worsened probability alignment for hard and intensity-weighted targets while improving it for uniform soft targets \citep{ohman2026quality}. Sparse annotation can still carry signal. The right response is not necessarily to discard the annotator but to keep the distinction between a negative judgment, an explicit no emotion response, and no response at all. This is the same distinction that \citet{webergenzel2024varierr} argue has to be made before label variation can be told apart from annotation error, and it has to be made in the data, not after aggregation.

A2 also shows why within-annotator co-occurrence cannot be read the same way for every annotator --- and why, if the quantity and quality of the annotation are too low, it may not be readable at all. With 13 secondary and 3 tertiary labels, an A2 co-occurrence matrix mostly visualizes scarcity. The absence of strong pairings is procedural evidence, but it is not the same kind of evidence as the dense A1 and A3 matrices, from which we can make more robust hypotheses about within-annotator co-occurrence.

GoEmotions offers a procedural comparison, though not a replication of A2's sparsity. Its annotators could mark an item as \texttt{example\_very\_unclear} instead of assigning emotion labels, and the raw data contain 3,411 such rows out of 211,225. Among raters with at least 100 annotations, the observed unclear rate ranges from 0 to approximately 12.6\%. An unclear response is neither a negative judgment about every emotion nor a missing annotation; it is an explicit statement that the annotator could not make the judgment. The comparison reinforces the distinction drawn above between a substantive negative judgment, an explicit uncertainty or abstention response, and no or unusually thin participation. Those states should not be collapsed before analysis.

\subsection{Presence and intensity disagreement are separable}
\label{ssec:presence}

\begin{figure}[t]
\centering
\includegraphics[width=0.95\linewidth]{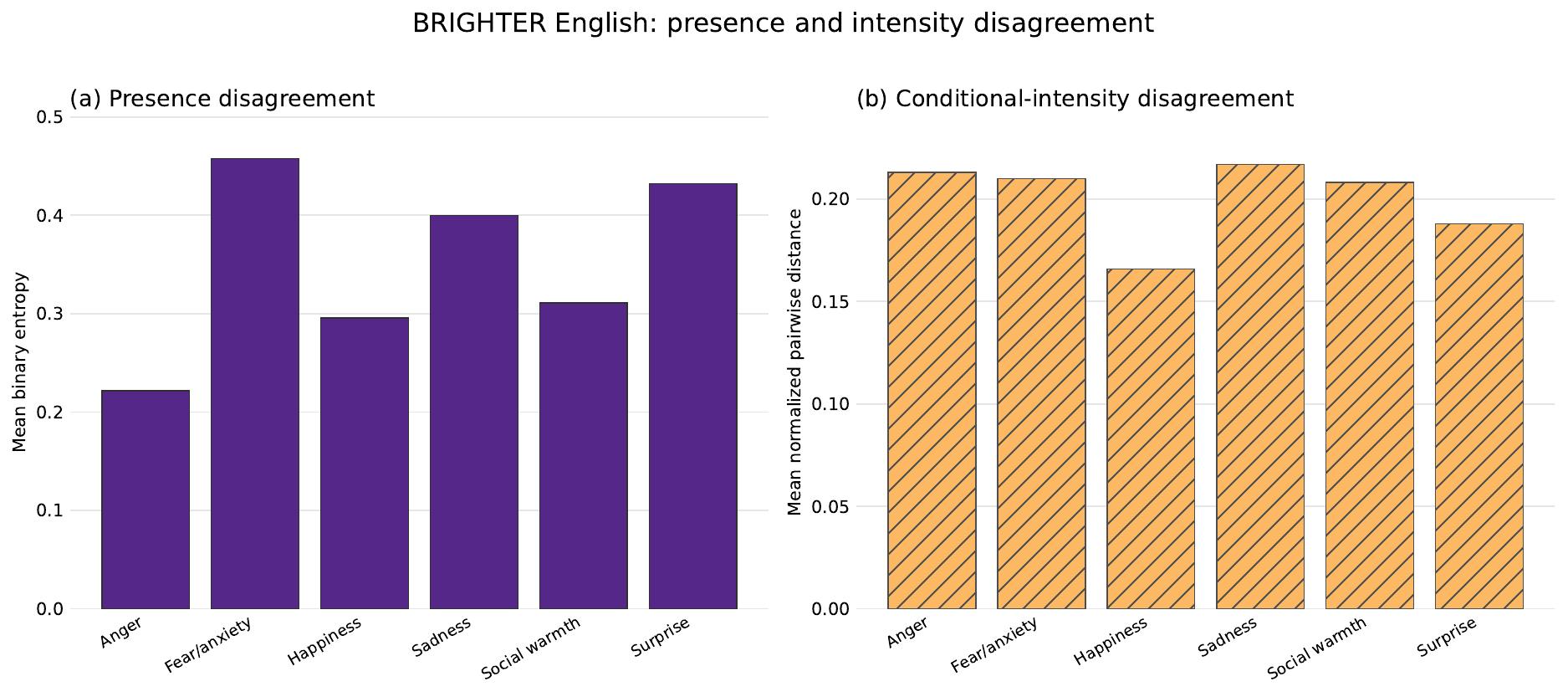}
\caption{BRIGHTER English disagreement by emotion category. Panel (a) shows mean disagreement over emotion presence, measured as binary entropy over ratings of 0 versus greater than 0. Panel (b) shows mean conditional-intensity disagreement among annotators assigning positive ratings, measured as normalized mean pairwise absolute distance and restricted to item--emotion combinations with at least three positive ratings. The panels use separate y-axes because the two measures are not commensurate in absolute magnitude.}
\label{fig:brighter-presence-intensity}
\end{figure}

The three TOMATS signatures concern which categories are chosen, together, by whom, or not at all. BRIGHTER allows a further question that TOMATS cannot answer cleanly: when annotators disagree about an emotion, do they disagree about whether it is there, or about how strong it is? Because BRIGHTER rates every emotion dimension on an ordinal scale, the two can be measured separately: presence disagreement from the split between ratings of 0 and ratings above 0, and conditional-intensity disagreement from the spread among positive ratings only, on items with at least three positive ratings (Figure~\ref{fig:brighter-presence-intensity}). This therefore is not a refinement of one of the TOMATS signatures but an additional dimension of disagreement that concerns the scale on which a recognized category is judged.

Presence disagreement varies markedly across emotions. Mean binary entropy ranges from 0.222 for anger to 0.457 for fear/anxiety, with surprise (0.432) and sadness (0.400) also showing relatively high disagreement. Conditional-intensity disagreement is much flatter across categories: mean normalized pairwise distance among positive ratings ranges from 0.166 for happiness to 0.217 for sadness, with the remaining emotions falling within this narrow interval. The two measures operationalize different aspects of disagreement and are not commensurate in absolute magnitude, so the important point is the difference in their profiles rather than a direct comparison of bar heights. Annotators can disagree over whether an emotion applies without substantially disagreeing over its intensity once recognized, and the relationship between these two forms of disagreement differs by emotion category.

\subsection{The signatures across datasets}
\label{ssec:synthesis}

Table~\ref{tab:6} summarizes the disagreement structure across the three resources. BRIGHTER independently shows structured compatibility between emotion categories. GoEmotions confirms general differences in annotation breadth and shows that pair-specific boundary behavior remains after breadth is controlled, separating two components that TOMATS cannot distinguish cleanly.

\begin{table*}[th!]
\centering
\caption{Summary of disagreement structure across TOMATS, BRIGHTER, and GoEmotions.}
\label{tab:6}
\small
\renewcommand{\arraystretch}{1.4}
\begin{tabularx}{\textwidth}{
>{\raggedright\arraybackslash}p{0.16\linewidth}
>{\raggedright\arraybackslash}X
>{\raggedright\arraybackslash}X
>{\raggedright\arraybackslash}X}
\toprule
Disagreement dimension & TOMATS & BRIGHTER & GoEmotions \\
\midrule

Structured category compatibility
& Recurrent cross-annotator alternatives (anger--disgust; fear--anticipation in one direction)
& Structured within-judgment lift; strongest for happiness--social warmth
& Not tested separately \\

General response style
& A3 averages 2.21 core labels per labelled item versus 2.06 for A1; not cleanly separable from boundary style
& Not identifiable without persistent rater IDs
& Item-adjusted cardinality ranges from $-0.235$ to $+1.936$ across 76 raters \\

Pair-specific boundary behavior
& Different within-annotator association patterns (e.g.\ anger--joy); not cleanly separable from general breadth
& Not identifiable without persistent rater IDs
& Merging/separating residuals remain after controlling for general breadth \\

Procedural sparsity/abstention
& A2 sparsity substantially affects apparent agreement
& Not available in the analyzed rating structure
& Explicit \texttt{very\_unclear} response; use varies substantially across raters \\

Presence vs.\ intensity
& Not cleanly separable given the original intensity instruction
& Directly separable; presence and conditional-intensity profiles differ by emotion
& No intensity ratings \\
\bottomrule
\end{tabularx}
\end{table*}

\section{What Aggregation Hides}

\subsection{Different processes, same representation}

The analyses above show that low agreement can arise from several distinct processes: recurring semantic overlap between categories, annotator-specific boundary placement, differences in annotation breadth, and procedural sparsity. Common aggregation methods collapse these processes to different degrees, because aggregation is many-to-one: distinct configurations of individual judgments can produce the same aggregate representation, as shown in Table~\ref{tab:4}.

\begin{table*}[t]
\centering
\caption{Different annotator configurations can produce identical marginal vote shares}
\label{tab:4}
\small
\renewcommand{\arraystretch}{1.4}
\begin{tabularx}{\textwidth}{>{\raggedright\arraybackslash}p{0.20\linewidth}>{\raggedright\arraybackslash}X>{\raggedright\arraybackslash}X>{\raggedright\arraybackslash}X>{\raggedright\arraybackslash}X>{\raggedright\arraybackslash}X}
\toprule
Configuration & A1 & A2 & A3 & Fear vote share & Anticipation vote share \\
\midrule
A: co-occurrence & fear + anticipation & fear & no emotion & 2/3 & 1/3 \\
B: competing readings & fear & fear & anticipation & 2/3 & 1/3 \\
C: sparse participation & fear + anticipation & fear & blank & 2/3* & 1/3* \\
\bottomrule
\multicolumn{6}{l}{\footnotesize *If the blank is treated as a negative response; excluding it from the denominator gives 1 and 1/2.}
\end{tabularx}
\end{table*}

A and B in Table~\ref{tab:4} produce identical marginals and mean different things. In A, anticipation is an extra emotion alongside fear in one annotator's judgment. In B, it is an alternative reading from a different annotator. The vote-share vector cannot tell them apart.
C adds a procedural difference. Treat the missing response as a negative label and it produces the same marginal vote shares as A. Track it as missing and the denominator changes, and so does the representation. Which interpretation applies is a fact about the annotation record, not about the resulting vector.

Hard aggregation loses more. Majority voting returns only fear in all three cases. A union-based hard target returns fear and anticipation in all three. Neither captures co-occurrence, disagreement between annotators, or missingness.

BRIGHTER supplies a second many-to-one case, on a different scale. Two items can have the same proportion of annotators who judge an emotion present and still differ in how those annotators rate its strength: in one, they may agree that the emotion is mild; in the other, they may split between mild and strong. Conversely, the same mean rating can come from broad agreement on a moderate intensity or from a split between annotators who see no emotion and annotators who see a strong one. We keep this out of Table~\ref{tab:4} because the measurement scale is different, but the point is the same.

Soft labels are a real improvement when the goal is to preserve uncertainty, but they do not solve the provenance problem on their own. They preserve marginal disagreement (how often each label was chosen) but not relational disagreement (which labels were chosen together, by whom, and against what response pattern). This is also why the choice of soft-label metric is contested: different measures emphasize different aspects of the distribution and none of them recovers the structure behind it \citep{rizzi2024soft}. A distribution over graded ratings presents similar issues. It keeps the spread of intensity judgments, but does not by itself say whether that spread comes mainly from disagreement over presence or from disagreement over strength once presence is granted. That is a second kind of information loss, beyond provenance: a soft target preserves uncertainty, but it is not, by itself, a diagnosis of what kind of uncertainty produced it.

\subsection{What to keep, and what to inspect}

Annotation projects should keep, wherever they can, (1) the original annotator-level labels; (2) explicit indicators separating missingness from negative or no-emotion judgments; (3) the original per-dimension ratings and persistent anonymized annotator identifiers; and (4) whatever aggregate the modeling requires. The aggregate then becomes a view of the data rather than the data itself. Level (3) is what made the response-style, boundary, and conditional-intensity analyses above possible. This is essentially the argument \citet{prabhakaran2021releasing} make about releasing annotator-level labels, and per-annotator modeling can match or beat aggregation while giving better-calibrated uncertainty \citep{davani2022dealing}.

This does not mean all disagreement is valuable. Careless annotation and misread instructions are real problems, and the A2 case is exactly why they have to be investigated rather than assumed either way. The signatures give a small vocabulary for that investigation: above-chance pairings point to category overlap or textual ambiguity; item-adjusted label cardinality to general annotation style, and pair-specific residuals to category boundaries; differences in coverage or abstention to procedural variation; and where ratings are graded, presence and conditional-intensity disagreement should be reported separately. All of these can be computed from annotator-level tables before aggregation.

None of this is specific to literary text. Structured overlap, response style, local boundaries, and procedural responses also shape large crowdsourced resources that are treated as model targets. Aggregated targets from such a resource are partly shaped by the response patterns of whoever judged each item, and models trained and evaluated against them learn to reproduce that aggregate. Low agreement or high label entropy is therefore an observation rather than a diagnosis.

\subsection{Implications for interpretive text research}

In literary studies, and in any discipline that deals with language data more broadly, ambiguity and vagueness are usually a feature of the object, not a defect in the measurement. The same sentence can carry different meanings for different people (ambiguity), or it may be unclear what precise meaning can be assigned to it (vagueness). With reference to emotions specifically, and as documented in previous research \citep{lange2023comparing,stella2022cognitive,diegoli2026shame}, emotions rarely, if ever, occur in isolation. Our diagnostics recognize that we normally feel multiple things at once, and this complexity is reflected in how annotators infer emotion from texts. A passage that one reader takes as fearful and another as anticipatory may be interesting precisely because the choice cannot be resolved: there is no `right' or `wrong', but rather a range of plausible readings.

The effect is not limited to individual passages. If annotators differ systematically in how they combine categories, an aggregated dataset can produce an emotional ontology that matches none of the readers who produced it. The final labels become a statistical compromise, an epiphenomenon of the aggregation process, detached from the structures through which any individual annotator reads the text.

Simplification is of course an unavoidable result of annotation, but not every simplification costs the same. Keeping provenance makes simplification reversible: we can aggregate for one purpose without permanently erasing the judgments the aggregate came from. 

Although disagreement in emotion annotation has often been studied from a cross-linguistic perspective, treating language and culture as major sources of variation in judgments, we make no such claim here. Perceptions of emotion can certainly vary across languages and cultures, but our study shows substantial variability even among members of the same in-group, for example between Annotators 1 and 3 in TOMATS. Comparable individual variability has also been observed in evaluative judgments in pragmatics \citep{haugh2019apology,baker2012individual}. All three TOMATS annotators were women in their mid-twenties, native English speakers, and affiliated with the same research group. They differed in upbringing and educational exposure, but three annotators cannot support claims about broader reader populations. What the results do show is why such claims require annotator-level data. If background, training, language, or cultural position affects where category boundaries fall, those differences disappear before analysis begins when only aggregated targets are retained. This is particularly notable given the relative homogeneity of the TOMATS annotator pool: substantial divergence emerges even under these conditions, suggesting that individual variability in emotion annotation needs to be operationalized in its own right.

These are good reasons to treat annotated humanities data as constructed and (to some degree) subjective representations rather than measurements \citep{drucker2011humanities}. Conceptualizing annotated data as representations does not eliminate the subjectivity, but makes more of the variation involved in categorization available for inspection. These issues are involved in a wide range of semantic/functional categorization procedures across disciplines. Within the neighboring disciplines of corpus linguistics and discourse analysis, a number of studies have discussed the implications of subjective annotation choices for the robustness of the interpretations \citep{fuoli2015optimising,fuoli2018stepwise,marchi2009winter}. The disagreement signatures addressed here are, however, rarely diagnosed jointly before aggregation. The practical tools we propose can facilitate more reliable, replicable and transparent protocols for many types of semantically oriented analysis, including stance, offensiveness, toxicity, moral judgment, and irony, which can all produce similar-looking disagreement distributions through different mechanisms.

\section{Conclusion}

Disagreement in emotion annotation takes several distinct forms. Across three resources with different designs, we find structured category overlap, differences in annotation breadth and in pair-specific category boundaries, procedural sparsity and abstention, and a separable distinction between disagreement over presence and over intensity.

These distinctions are hard to recover after aggregation. Hard labels suppress most of the variation. Soft vote-share labels keep uncertainty but lose the relational and procedural structure behind it, so two very different annotation sets can produce the same distribution. A soft target over graded ratings preserves uncertainty as well, but it is not, by itself, a diagnosis of whether that uncertainty concerns presence or strength.

The practical implication is simple: analyze disagreement before collapsing it, and retain annotator-level provenance whenever an aggregate is needed for modeling. This keeps variation in human judgments available for later analysis instead of treating it as residue. The same principle applies to other semantic annotation tasks in which plausible judgments may differ across annotators. We hope that this work contributes to a broader discussion of how such variation should be diagnosed, documented, and preserved.

\section*{Limitations}
\label{sec:limitations}

The typology is a starting point, not an exhaustive taxonomy. TOMATS is small: one English-language literary text and three annotators, one of them extremely sparse. BRIGHTER and GoEmotions broaden the evidence, but they are secondary observational analyses rather than controlled replications, and we do not compare absolute disagreement levels across the three resources.

The signatures are not experimentally identified mechanisms. Semantic overlap may come from the text, the taxonomy, or both; within-annotator co-occurrence may reflect stable boundaries, an annotation strategy, or priming; and the filtered agreement analysis confounds missingness with item selection. Residualizing GoEmotions boundary behavior controls for general breadth but does not show that any rater holds a stable internal emotion ontology. Separating these would require reannotation under randomized order, controlled manipulation of context and label presentation, and more annotators. Plutchik's eight categories are also a constraint: a different inventory or a dimensional representation would produce different signatures, which BRIGHTER and GoEmotions only partly address.

Two parts of the TOMATS annotation are set aside. The intensity field asks for an overall judgment of the passage and cannot be tied to individual emotions, although intensity-weighted soft targets aligned best with annotator uncertainty in the earlier study \citep{ohman2026quality}. The free-text ``other'' field is the larger omission. A3 in particular used it: for one item A1 gave anticipation and fear while A3 left the primary slot empty and wrote pessimism, arguably a combination of the two. Several free-text labels (contempt, submission, sentimentality, unbelief, hope, remorse) are Plutchik dyad names, which suggests the annotators were reasoning compositionally in a way the interface could not record. Out-of-taxonomy annotation may therefore be a further disagreement signature and deserves separate analysis.

\section*{Data Availability Statement}

The processing code and derived annotation tables used in this study are available through the anonymous repository at \url{https://github.com/esohman/DisagreementTypes}. For TOMATS, the released annotation and derived tables are keyed by sentence index and do not redistribute the copyrighted text of \textit{The Old Man and the Sea}; the analyses can be reproduced by applying the released code to a legally obtained copy of the novel. BRIGHTER and GoEmotions are available from their original distributions \citep{muhammad2025brighter,demszky2020goemotions} and are not redistributed here; we release only analysis code and derived summary outputs, which can be regenerated from the source data. All three datasets are distributed under CC BY 4.0 licenses. This article reanalyzes TOMATS, BRIGHTER, and GoEmotions and introduces no new participant data. The analyses are computationally lightweight and can be reproduced on a standard personal computer without specialized hardware.

\section*{Acknowledgements}

This work was supported by JSPS KAKENHI Grant Number 24K21058 and by the Next-GenerationEU National Recovery and Resilience Plan (NRRP) -- SELFEE project, CUP N.H73C24001500001.

Computational resources were provided by CSC -- IT Center for Science, Finland, through the Roihu computing environment.

Claude Code was used to consolidate and tidy up the code for the repository.

\bibliographystyle{compling}
\bibliography{disagreement,COLI_template}

\end{document}